\documentclass[review]{elsarticle}

\usepackage{lineno,hyperref}

\journal{Neurocomputing}
\usepackage{amsmath,amsmath,amssymb}
\usepackage{multirow}

\begin{document}

\begin{frontmatter}

\title{Cascade Network for Self-Supervised Monocular Depth Estimation}


\author[mymainaddress]{Chunlai Chai}

\author[mymainaddress]{Yukuan Lou\corref{mycorrespondingauthor}}
\cortext[mycorrespondingauthor]{Corresponding author}
\author[mymainaddress]{Shijin Zhang}

\address[mymainaddress]{School of computer and information engineering, Zhejiang Gongshang University, Hangzhou, 310018, China}

\begin{abstract}
It is a classical compute vision problem to obtain real scene depth maps by using a monocular camera, which has been widely concerned in recent years. However, training this model usually requires a large number of artificially labeled samples. To solve this problem, some researchers use a self-supervised learning model to overcome this problem and reduce the dependence on manually labeled data. Nevertheless, the accuracy and reliability of these methods have not reached the expected standard. In this paper, we propose a new self-supervised learning method based on cascade networks. Compared with the previous self-supervised methods, our method has improved accuracy and reliability, and we have proved this by experiments. We show a cascaded neural network that divides the target scene into parts of different sight distances and trains them separately to generate a better depth map. Our approach is divided into the following four steps. In the first step, we use the self-supervised model to estimate the depth of the scene roughly. In the second step, the depth of the scene generated in the first step is used as a label to divide the scene into different depth parts. The third step is to use models with different parameters to generate depth maps of different depth parts in the target scene, and the fourth step is to fuse the depth map. Through the ablation study, we demonstrated the effectiveness of each component individually and showed high-quality, state-of-the-art results in the KITTI benchmark.
\end{abstract}

\begin{keyword}
Cascade Network, Monocular Depth Estimation, Self-supervision
\end{keyword}

\end{frontmatter}


\section{Introduction}
Obtaining a depth map of the target scene is a classic computer vision problem, and it is widely used in automatic driving \cite{1}, 3D reconstruction \cite{2}, and other fields. Inaccurate depth maps will bring a negative impact on the algorithms in these fields. Therefore, how to generate a more accurate and reliable depth map has become an urgent need.

In recent years, the generation of scene depth map has become a hot topic in computer vision research. At present, the generation of the scene depth map is mainly divided into two categories. The first one is to obtain depth information directly through hardware devices, such as SLAM \cite{3}, Kinect \cite{4}, etc. The advantage of this method is that the depth information can be obtained accurately and quickly. However, the disadvantage of this method is that it requires additional hardware costs. The second method is to restore depth information from a two-dimensional image by computer vision technology.

In this paper, due to the advantages of deep learning in feature representation and generalization ability, we focus on using deep learning to generate monocular depth maps. However, deep learning relies on a large number of artificially labeled samples. The lack of data sets can cause the model to perform poorly when generating depth maps of complex or rare scenes. The self-supervised / unsupervised model can solve this problem. Therefore, various models based on self-supervised / unsupervised have emerged recently. For example, Zhou et al. \cite{5} proposed two independent models, one for monocular depth prediction and the other for multi-view camera pose estimation; the two models cooperate to complete self-supervised learning. After that, Yin et al. \cite{6} introduced optical flow estimation in the model and divided the target scene into a static background and a dynamic target, generating a more accurate depth map. Recently, Godard et al. \cite{7} have proposed a new multi-scale network to generate depth maps and obtain better results by using new constraints and auto-mask approach.

\begin{figure}[t]
	\centering
	\includegraphics[height=6.3cm,width=12cm]{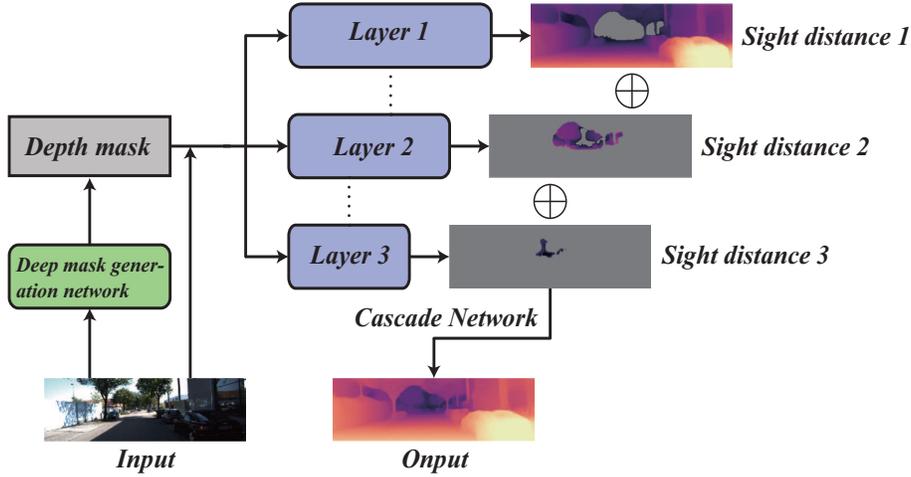}
	\caption{Overview of our framework. The whole framework is divided into different parts according to the different sight distances, and the scene with different sight distances is predicted, respectively. Finally, the depth map of fusion is output.}
	\label{figl}
\end{figure}

Although self-supervised learning has made remarkable achievements in depth estimation, there are still some limitations to these methods. These methods do not fully consider the impact of different sight distance scenes on the model's accuracy and use the one model to deal with the parts with different sight distances in the one scene. It is limiting the accuracy of a depth map. Especially the further scene, the depth map generated is often very fuzzy.

Based on the above analysis, we propose a model called cascade monocular depth estimation network (CMDEN), which is a cascaded neural network based on self-supervised learning, as shown in Figure 1. This model can deal with different sight distance parts separately in one scene. In particular, the auto-mask approach can be very good at shielding the different sight distance parts of the scene, so that each layer of the cascade network does not interfere with each other. They learn the depth features within a specific sight distance of the scene and generate an accurate depth map. Finally, the depth maps of each part are fused to output the final scene depth map.

The contributions of this study are summarized as follows:
\begin{enumerate}[(1)]
	\item We design a new cascade network based on self-supervised learning, which can process different sight distance parts of the one scene.
	\item We propose a new and straightforward method to shield pixels at different sight distances to reduce the impact of scenes at different sight distances on model training.
	\item Our model achieved advanced results in the KITTI benchmark.
\end{enumerate}
%
%

\section{Related Works}
In this section, we review models that use monocular color image input and predict each pixel's depth as the output.
 
\subsection{Supervised Depth Estimation}
The depth estimation task is to predict from two-dimensional image to depth map, so the whole process is a self-coding process, including encoding and decoding. In the early monocular depth estimation network framework, the whole problem was treated as a regression problem, which directly used the above-mentioned basic framework, and adopted a multi-scale and deeper network model \cite{8}. The disadvantage of turning the depth estimation problem into a regression problem is that it depends too much on the data set. So some researchers have turned the depth estimation into a classification problem, dividing the nearest distance to the farthest actual distance into several different categories \cite{9}. 

Some researchers regard the depth map as an image style and introduce image style transfer and generative adversarial networks \cite{42} into the monocular depth estimation field \cite{10,11}.
\subsection{Self-supervised/Unsupervised Depth Estimation}
The supervised algorithm is limited by data set. Therefore, many self-supervised/unsupervised algorithms have been proposed in recent years, including the algorithm based on stereo pairs. For example, the depth estimation problem is transformed into the stereo matching problem of the left and right views. The model first predicts another view's disparity according to the original image and then outputs the depth map combined with the other view \cite{12,13}. 

Depth estimation algorithm based on monocular data. A less constrained form of self-supervised/unsupervised is to use consecutive temporal frames to provide training signals. In addition to predicting the depth of the scene, researchers must also design a network to estimate the camera's real-time pose, which is a challenge because there are moving objects in the scene. The training of depth network needs the constraint of camera pose. 

Generally speaking, the accuracy of monocular depth estimation lags behind stereo pairs matching, but recent methods have begun to narrow the performance gap between them. Researchers have proposed a variety of methods, such as restricting the consistency between the predicted depth and the predicted surface normal \cite{14} and enhancing the consistency of the edge \cite{15}. A matching loss based on approximate geometry is proposed to promote temporal depth consistency \cite{16}. The depth normalization layer is used to overcome the preference for smaller depth values \cite{17}, which comes from the depth smoothing term commonly used \cite{12}. Use pre-computed instances segmentation masks to help deal with moving objects \cite{18}.

Depth estimation algorithm based on video sequences. For example, the motion generated by adjacent video frames is used to approximate the multi-view image, and the camera pose can be estimated, so as to complete the SLAM work further \cite{5,19}. After that, some researchers added optical flow information \cite{6} to generate more accurate depth maps. 

Depth estimation algorithm based on multi-tasking. For example, the multi-task framework based on the image segmentation algorithm RefineNet \cite{20} can perform depth estimation and target segmentation simultaneously. The combination of monocular depth estimation and optical flow prediction as a joint task framework \cite{21}, which is different from training two tasks separately, the author considers the consistency of the two tasks to promote each other. Besides, some researchers have proposed a multi-task integration framework that integrates monocular depth estimation, camera pose estimation, optical flow estimation, and motion segmentation \cite{22}.

\begin{figure}[ht]
	\centering
	\includegraphics[height=11.6cm,width=11cm]{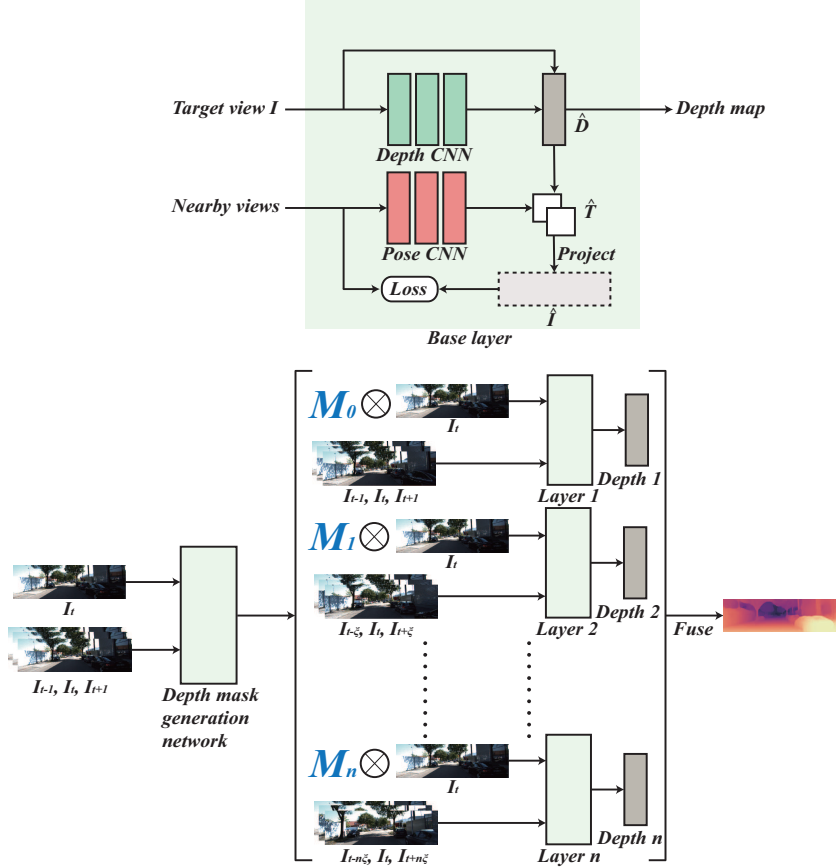}
	\caption{Overview of the self-supervised learning models based on cascaded networks. This model is divided into three stages. In the first stage, we use the base layer as the depth mask generation network. The network outputs masks with different sight distances and divides the scene into several parts. In the second stage, we use the base layer to process every part of the target scene. In the third stage, we fuse the depth maps of each part as a result. $\xi$ denotes the interval between two frames.}
	\label{fig2}
\end{figure}

\section{Method}
Here we propose a monocular depth estimation network based on the cascade strategy and a camera pose estimation network from unlabeled video sequences, as shown in Figure 2. The depth estimation network and pose estimation network are optimized together in training and used separately in the test. The number of layers of the cascade network can be adjusted according to the scene to obtain more accurate results. The data set used in training is the video sequence captured by the moving camera. It is assumed that most of our target scenes are rigid; that is, the scene changes between different frames are dominated by camera motion.

\subsection{Notation and Problem Formulation}
We will specify the mathematical notation as follows: Use $<\emph{I}_1,...,\emph{I}_N>$ denotes a continuous image sequence in the training set, where $\emph{I}_t$ denotes the target view, and $\emph{I}_s(1<s<N, s\ne t)$ denotes the source views. Use $\emph{p}$ denotes the homogeneous coordinates of a pixel in the view. Use $\hat{D}_t$ and $\hat{T}_{t\to s}$ to denote the predicted depth map and camera pose, respectively. Use $\emph{K}$ denotes the known camera intrinsics matrix.

\subsection{Base Layer}
This section first reviews the key ideas behind self-supervised training for monocular depth estimation and then describes the basic layer in our model. 

As described in \cite{14}, the key to applying self-supervised learning to monocular depth estimation is to use the depth estimation network and the pose estimation network to generate the depth and camera pose of the target view, respectively, and then use the depth map and camera pose to project the source view into a synthesize view that is as similar as possible to the target view. Moreover, the whole process can be achieved in a fully differentiable way using a convolution network. In our model, we encapsulate the above steps as a base layer. The whole projection relation can be expressed as
\begin{equation}\label{1}
\emph{p}_s \sim \emph{K}\hat{T}_{t\to s}\hat{D}_t(\emph{p}_t)\emph{K}^{-1}\emph{p}_t
\end{equation}
or
\begin{equation}\label{2}
\hat{I}_t=\emph{proj}(\hat{T}_{t\to s},\hat{D}_t(\emph{I}_t),\emph{K},\emph{I}_s)
\end{equation}
where $\emph{p}_t$ denotes a pixel in the target view, $\emph{p}_s$ denotes a pixel in the source view, and $\hat{I}_t$ denotes synthetic target view. $\emph{proj}()$ represents the operation of projection. For notation simplicity, we omit showing the necessary conversion to homogeneous coordinates along the steps of matrix multiplication.

Similar to \cite{7}, we also optimize our model
as the minimization of a photometric reprojection error at training time. When we get $\hat{I}_t$, we optimize the model by
\begin{equation}\label{3}
\mathcal{L}_{p}=\sum_{s}\emph{pe}(\emph{I}_t,\hat{I}_t)
\end{equation}
where $\emph{pe}()$ represents the photometric reprojection error, and we follow \cite{23,12} in using $\emph{L1}$ and $\emph{SSIM}$ \cite{24} to make our photometric error function $\emph{pe}()$, i.e.
\begin{equation}\label{4}
	\emph{pe}(\emph{I}_a,\emph{I}_b)=\frac{\alpha}{2}(1-SSIM(\emph{I}_a,\emph{I}_b))+(1-\alpha)||\emph{I}_a-\emph{I}_b||_{1}
\end{equation}
\begin{equation}\label{5}
\emph{SSIM}(\emph{x},\emph{y})=\dfrac{(2\mu_x\mu_y+\emph{c}_1)(2\sigma_{xy}+\emph{c}_2)}{(\mu_x^2+\mu_y^2+\emph{c}_1)(\sigma_x^2+\sigma_y^2+\emph{c}_2)}
\end{equation}
where $\alpha=0.85$, $\mu$ is the mean value of the image, $\sigma$ is the variance or covariance of the image, and $\emph{c}$ is a constant to prevent the equation from dividing by zero.

As in \cite{12} we use edge-aware smoothness to discourage shrinking of the estimated depth
\begin{equation}\label{6}
\mathcal{L}_s=|\partial_xd_t^*|\emph{e}^{-|\partial_x\emph{I}_t|}+|\partial_yd_t^*|\emph{e}^{-|\partial_y\emph{I}_t|}
\end{equation}
where $d_t^*=d_t/\overline{d_t}$ is the mean-normalized inverse depth from \cite{25}. 

Since self-supervised monocular training usually operates under the assumption of moving cameras and static scenes, the performance will be significantly affected when objects are moving. To reduce the impact of moving objects on model training, we use a simple auto-masking method \cite{7} to mask the pixels moving between two frames, As shown in the following formula
\begin{equation}\label{7}
\mu=[\mathop{min}\limits_{s}\emph{pe}(\emph{I}_t,\hat{I}_t)<\mathop{min}\limits_{s}\emph{pe}(\emph{I}_t,\emph{I}_s)]
\end{equation}	
This operation is only used in the depth mask generation network in our model and is not used in the subsequent stages.

To prevent the training objective of getting stuck in local minima \cite{43}, the existing model uses multi-scale depth prediction and image reconstruction \cite{44,12}. In this way, the total loss is a combination of the decoder's mid-layer loss. However, as described in \cite{7},  the low-texture area of low-resolution mid-layer depth map tends to create "holes",  so inspired by \cite{45}, we up-sample the low-resolution image of the mid-layer to the input image resolution and then calculate the loss. This effectively constrains the depth maps at each scale to work toward the same objective i.e. reconstructing the high resolution input target image as accurately as possible.

We combine the above two loss functions, $\mathcal{L}_p$ and $\mathcal{L}_s$, as the final loss function $\mathcal{L}=\mu\mathcal{L}_p+\lambda\mathcal{L}_s$, and average over each pixel, scale and batch.

\begin{figure}[ht]
	\centering
	\includegraphics[height=5.06cm,width=11.25cm]{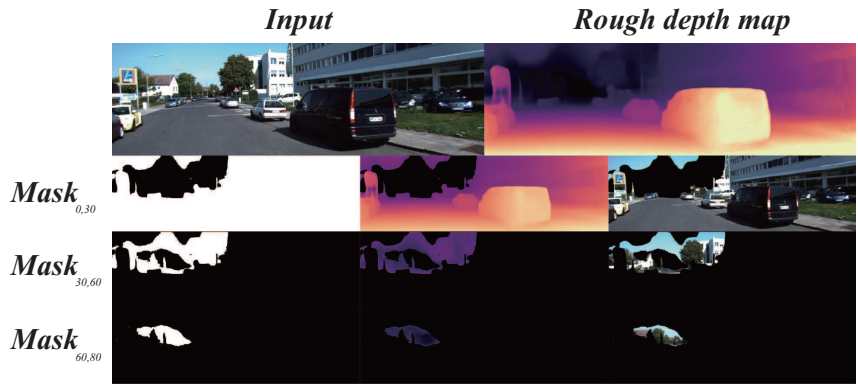}
	\caption{An overview of depth mask. The first line is the target scene image, and a rough depth map is used as a label that divides the sight distance. The next few lines show the scene and depth mask divided by different sight distance.}
	\label{fig3}
\end{figure}

\subsection{Depth Mask Generation Network}
As described in Section 3.2, we used the base layer as the depth mask generation network and generated a rough depth map. Then, the generated rough depth map was used as the label to divide the sight distance. As shown in Figure 3. The target scene was divided into several parts with a different sight distance using the following formula
\begin{equation}\label{8}
\emph{mask}_{\alpha,\beta}[i][j]=\begin{cases}
1,\quad \alpha\le\hat{D}_t(\emph{I}_t)[i][j]<\beta\\
0,\quad \emph{others}
\end{cases}, 0\le \alpha<\beta\le \emph{MAX}(\hat{D}_t(\emph{I}_t))
\end{equation}
where $\alpha$ and $\beta$ denote the mask's sight distance, and $\emph{MAX}()$ denotes the maximum depth of the estimated depth map.

\begin{figure}[ht]
	\centering
	\includegraphics[height=7.5cm,width=10cm]{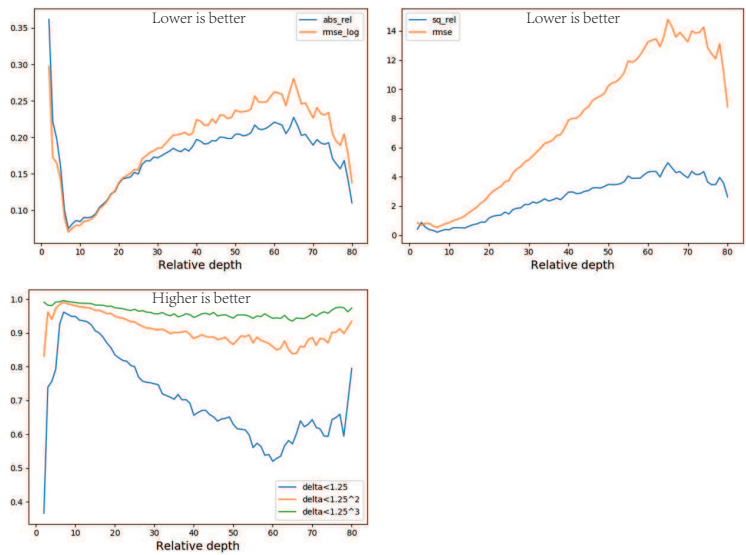}
	\caption{Accuracy statistics for rough depth maps, the x-axis is relative depth, and different polylines represent relative error, root mean square error, logarithmic error, and correct, respectively.}
	\label{fig4}
\end{figure}

\subsection{Cascade Network}
Before we begin this section, let us first answer why we use cascade networks to learn about different sight distance parts in a scene. First, we evaluate the accuracy of the rough depth map. As shown in Figure 4, we can find two problems: one is that the accuracy of the model is very low when the sight distance is very close; the other is that the accuracy of the depth map decreases with the increase of the sight distance.

For the first problem, we find that, as shown in equation 1, the model is optimized by projecting the source view to target view. Because view changes between different frames are dominated by camera motion, the loss of close sight distance scenes can occur between two frames. As shown in the first row of Figure 5, part of the scene in this frame is missing in the next frame, which leads to low accuracy of the close sight distance scene when the model is optimized. Nevertheless, to some extent, this is a problem caused by self-supervised learning. The cascade network we are proposing is to solve the second problem. As the sight distance increases, the parallax between the target and source views decreases. As shown in the second line of Figure 5, the parallax of two frames at a close sight distance is much greater than that at a far sight distance, even the parallax of two adjacent frames in the far sight distance is almost zero. It is difficult for a model to learn by projecting one view to another view with little or almost no parallax; this is why the depth map's accuracy decreases as the sight distance increases. However, suppose we project a far sight distance part of the scene to a scene with sufficient parallax, or further, to project different sight distance parts of the target scene to different interval source scenes, so there will always be enough parallax for the model to learn, as shown in the third row of Figure 5. To ensure that the model is not interfered with by other sight distance parts when optimizing, we use the depth mask mentioned in the previous section to shield the other sight distance parts of the scene.

The whole process is shown in Figure 2. The training part of each layer of the cascade network still uses baselayer. After obtaining the depth map of each part, use the following formula to merge the parts
\begin{equation}\label{9}
\hat{D}_t(\emph{I}_t)=\sum_{i=1}^{n}\emph{mask}_{\alpha_i,\beta_i}\hat{D}_t(\emph{I}_t)_i
\end{equation}
where $n$ denotes the number of layers of the cascade network.

\begin{figure}[]
	\centering
	\includegraphics[height=9cm,width=12cm]{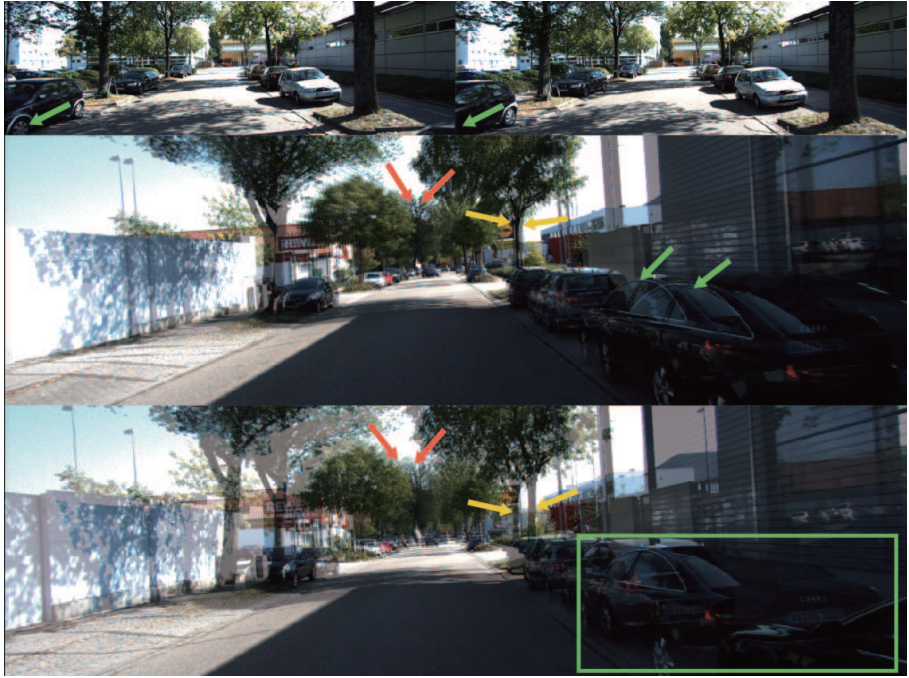}
	\caption{Some illustrations. The first line shows the problem of missing scenes in close sight distance, such as the missing tires in the picture. The second line shows that the parallax decreases as the sight distance increases. For example, from a car to a tree trunk to further branches and leaves, the parallax becomes smaller and smaller. The third line indicates that the more interval frames can ensure that the further scenes still have enough parallax. The disadvantage is that the close sight distance scenes will be completely missing, such as the disappearing Audi car in the lower right corner of the picture. So we propose a depth mask to avoid the interference of these scenes on model optimization.}
	\label{fig5}
\end{figure}

\subsection{Network architecture}
Our model's depth estimation network is based on the general U-NET structure \cite{26}, including decoders, encoders, and skip connections, which allows the depth estimation network to learn local features and global features simultaneously. The encoder uses ResNet18 \cite{27}, and the parameters used are pre-trained by ImageNet \cite{28}. The structure of the decoder is similar to \cite{12}, we convert the output $\sigma$ to depth with $\emph{D}=1/(a\sigma+b)$, where $a$ and $b$ are chosen to constrain $\emph{D}$ between 0.1 and 100 units. In the convolution operation in the decoder, we use reflection padding, which can reduce border artifacts \cite{12}.

For the pose estimation network, we use ResNet18, modified to accept a pair of color images as input, and predict a single 6-DoF relative pose. Similar to \cite{7}, we perform horizontal flips and the following training augmentations, with 50$\%$ chance: random brightness, contrast, saturation, and hue jitter with respective ranges of $\pm$0.2, $\pm$0.2, $\pm$0.2, and $\pm$0.1. Only perform color enhancement processing on images input to the network.

\section{Experiments}
We evaluate our models, named CMDEN, on the KITTI 2015 stereo dataset \cite{29}, to allow comparison with previously published monocular methods.
\subsection{Dataset}
KITTI is currently the world's largest self-driving algorithm evaluation data set. This data set evaluates computer vision technologies such as stereoscopic images, optical flow, visual odometry, 3D object detection, and 3D tracking. KITTI contains real image data collected in scenes such as urban areas, rural areas, and highways.

We use the data split of Eigen et al. \cite{30}. When using monocular sequences for training, we use the method of Zhou et al. \cite{5} for pre-processing to remove static frames. There are 39,810 monocular triplets for training and 4,424 for verification in the entire data set. We use the same intrinsics for all images, set the camera's principal point as the image center, and set the focal length to the average of all focal lengths in KITTI. During the evaluation process, we controlled the depth to 80 meters according to standard practices \cite{12}. 
For our monocular models, we report results using the per-image median ground truth scaling introduced by \cite{5}.

\begin{table}[]
	\centering
	\caption{Quantitative results. We show our model's results under different experimental conditions, experimenting with two resolution images and four different frame intervals. The benchmark set of the experiment is KITTI. The best results for each category are shown in bold.}
	\setlength{\tabcolsep}{1mm}{
		\begin{tabular}{|l|c|c|c|c|c|c|c|c|}
			\hline
			\multicolumn{1}{|c|}{\multirow{2}{*}{Method}}                                                                                                                          & \multirow{2}{*}{Resolution}                                                                                                            & \multicolumn{4}{c|}{Lower is better}                                                                                                                                                                                                                                                                                                                                                                                                                         & \multicolumn{3}{c|}{Higher is better}                                                                                                                                                                                                                                                                                                         \\ \cline{3-9} 
			\multicolumn{1}{|c|}{}                                                                                                                                                 &                                                                                                                                        & Abs Rel                                                                                                      & Sq Rel                                                                                                        & RMSE                                                                                                          & RMSE log                                                                                                      & $\delta$\textless{}$1.25$                                                                                              & $\delta$\textless{}$1.25^2$                                                                                               & $\delta$\textless{}$1.25^3$                                                                                               \\ \hline
			\begin{tabular}[c]{@{}l@{}}Zhou \cite{5}\\ Mahjourian \cite{16}\\ Yin and Shi \cite{6}\\ Wang \cite{34}\\ Casser \cite{35,36}\\ Meng \cite{37}\\ Godard \cite{7}\\ Klingner \cite{38}\end{tabular} & \begin{tabular}[c]{@{}c@{}}416x128\\ 416x128\\ 416x128\\ 416x128\\ 416x128\\ 416x128\\ 416x128\\ 416x128\end{tabular}                  & \begin{tabular}[c]{@{}c@{}}0.198\\ 0.159\\ 0.153\\ 0.148\\ 0.141\\ 0.139\\ 0.128\\ 0.128\end{tabular}        & \begin{tabular}[c]{@{}c@{}}1.836\\ 1.231\\ 1.328\\ 1.187\\ 1.026\\ 0.949\\ 1.087\\ 1.003\end{tabular}         & \begin{tabular}[c]{@{}c@{}}6.565\\ 5.912\\ 5.737\\ 5.583\\ 5.291\\ 5.227\\ 5.171\\ 5.085\end{tabular}         & \begin{tabular}[c]{@{}c@{}}0.275\\ 0.243\\ 0.232\\ 0.228\\ 0.215\\ 0.214\\ 0.204\\ 0.206\end{tabular}         & \begin{tabular}[c]{@{}c@{}}0.718\\ 0.784\\ 0.802\\ 0.810\\ 0.816\\ 0.818\\ 0.855\\ 0.853\end{tabular}         & \begin{tabular}[c]{@{}c@{}}0.901\\ 0.923\\ 0.934\\ 0.936\\ 0.945\\ 0.945\\ 0.953\\ 0.951\end{tabular}         & \begin{tabular}[c]{@{}c@{}}0.960\\ 0.970\\ 0.972\\ 0.975\\ 0.979\\ 0.980\\ 0.978\\ 0.978\end{tabular}         \\ \hline
			\begin{tabular}[c]{@{}l@{}}Guizilini \cite{33}\\ Godard \cite{7}\\ Klingner \cite{38}\\ \textbf{CMDEN 2}\\ \textbf{CMDEN 4}\\ \textbf{CMDEN 6}\\ \textbf{CMDEN 8}\end{tabular}                                     & \begin{tabular}[c]{@{}c@{}}640x192\\ 640x192\\ 640x192\\ 640x192\\ 640x192\\ 640x192\\ 640x192\end{tabular}                            & \begin{tabular}[c]{@{}c@{}}0.117\\ 0.115\\ 0.117\\ \textbf{0.112}\\ 0.113\\ 0.113\\ 0.115\end{tabular}                & \begin{tabular}[c]{@{}c@{}}0.854\\ 0.903\\ 0.907\\ \textbf{0.809}\\ 0.814\\ 0.824\\ 0.856\end{tabular}                 & \begin{tabular}[c]{@{}c@{}}\textbf{4.714}\\ 4.863\\ 4.844\\ 4.72\\ 4.78\\ 4.809\\ 4.94\end{tabular}                    & \begin{tabular}[c]{@{}c@{}}\textbf{0.191}\\ 0.193\\ 0.196\\ \textbf{0.191}\\ 0.192\\ 0.193\\ 0.197\end{tabular}                 & \begin{tabular}[c]{@{}c@{}}0.873\\ 0.877\\ 0.875\\ \textbf{0.878}\\ 0.877\\ 0.875\\ 0.874\end{tabular}                 & \begin{tabular}[c]{@{}c@{}}\textbf{0.963}\\ 0.959\\ 0.958\\ 0.960\\ 0.959\\ 0.959\\ 0.958\end{tabular}                 & \begin{tabular}[c]{@{}c@{}}\textbf{0.981}\\ \textbf{0.981}\\ 0.980\\ \textbf{0.981}\\\textbf{0.981}\\ \textbf{0.981}\\ \textbf{0.981}\end{tabular}                 \\ \hline
			\begin{tabular}[c]{@{}l@{}}Luo \cite{39}\\ Ranjan \cite{40}\\ Zhou \cite{41}\\ Klingner \cite{38}\\ Godard \cite{7}\\ \textbf{CMDEN 2}\\ \textbf{CMDEN 4}\\ \textbf{CMDEN 6}\\ \textbf{CMDEN 8}\end{tabular}            & \begin{tabular}[c]{@{}c@{}}832x256\\ 832x256\\ 1248x384\\ 1280x384\\ 1024x320\\ 1024x320\\ 1024x320\\ 1024x320\\ 1024x320\end{tabular} & \begin{tabular}[c]{@{}c@{}}0.140\\ 0.139\\ 0.121\\ 0.113\\ 0.115\\ \textbf{0.110}\\ 0.113\\ 0.114\\ 0.115\end{tabular} & \begin{tabular}[c]{@{}c@{}}1.029\\ 1.032\\ 0.837\\ 0.880\\ 0.882\\ \textbf{0.794}\\ 0.800\\ 0.802\\ 0.830\end{tabular} & \begin{tabular}[c]{@{}c@{}}5.350\\ 5.199\\ 4.945\\ 4.695\\ 4.701\\ \textbf{4.591}\\ 4.628\\ 4.645\\ 4.761\end{tabular} & \begin{tabular}[c]{@{}c@{}}0.216\\ 0.213\\ 0.197\\ 0.192\\ 0.190\\ \textbf{0.188}\\ 0.190\\ 0.190\\ 0.192\end{tabular} & \begin{tabular}[c]{@{}c@{}}0.816\\ 0.827\\ 0.853\\ \textbf{0.884}\\ 0.879\\ 0.879\\ 0.879\\ 0.877\\ 0.877\end{tabular} & \begin{tabular}[c]{@{}c@{}}0.941\\ 0.943\\ 0.955\\ 0.961\\ 0.961\\\textbf{0.962}\\ 0.961\\ 0.961\\ 0.960\end{tabular} & \begin{tabular}[c]{@{}c@{}}0.976\\ 0.977\\ \textbf{0.982}\\ 0.981\\ \textbf{0.982}\\ \textbf{0.982}\\ \textbf{0.982}\\ \textbf{0.982}\\ 0.981\end{tabular} \\ \hline
	\end{tabular}}
\end{table}

\subsection{Experimental Settings}
We implemented our framework with Pytorch \cite{31}. During the training process, we used Adam \cite{32} to optimize the model. A total of 20 epochs were trained. The batch size is 6, and the image resolution used is 1024x320/640x192. In the first 15 epochs, we used a learning rate of ${10}^{-4}$, and we reduce the learning rate to ${10}^{-5}$ in the last 5 epochs. The smoothness term $\lambda$ is set to 0.001. To simplify the results, we use a three-layer cascade network with the relative depth of each layer being $[0, 30)$, $[30, 60)$, and $[60, 80)$. The interval $\xi$ between two frames is 2/4/6/8.

\subsection{Monocular depth estimation}
We show the different quantitative results of our model under different experimental conditions, as shown in Table 1. It can be seen that compared to low resolution, our model performs better at high resolution. In the comparison of different frame intervals, interval 2 performs best, i.e., the first layer has a frame interval of 1, and the second Layer interval 2, third layer interval 4. In the comparison of different evaluation indicators, we can find that our model is not very sensitive to the accuracy indicator, and the change of frame interval has no significant influence on it; it is relatively sensitive to relative error and root mean squared error,  the model performs well in a reasonable range of frame interval. The results show that our monocular method outperforms existing state-of-the-art self-supervised approaches. The qualitative results can be seen in Figure 6.

\begin{figure}[ht]
	\centering
	\includegraphics[height=12cm,width=10cm]{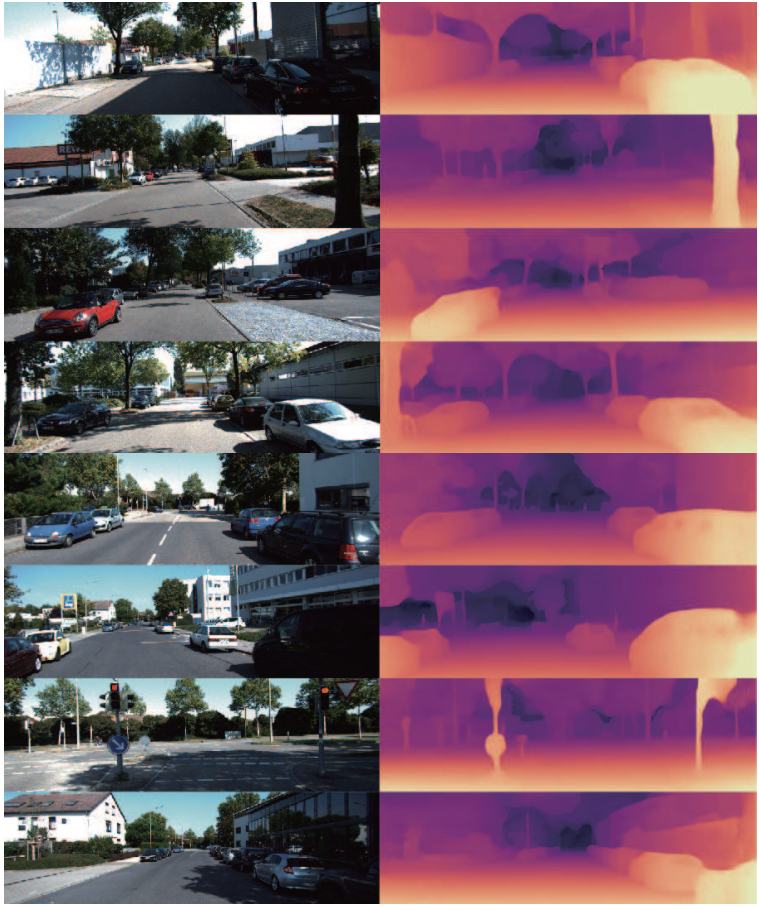}
	\caption{Qualitative results. Our model can generate a reasonably clear depth map, which is reflected in the superior quantitative results in Table 1.}
	\label{fig6}
\end{figure}

\subsection{Ablation Study}
To better understand how our cascade network affects the overall performance of monocular training, as shown in Table 2, we conduct ablation studies by changing the number of layers of the cascade network. We see that a cascaded network with only one layer, i.e., the baseline, performs the worst. When the number of network layers increases, the results will be significantly improved. 

\begin{table}[ht]
	\centering
	\caption{Ablation study. We show the influence of cascaded networks with different layers on the results. The baseline does not use a cascaded structure, and interval $\lambda$ between two frames is 2. The experiment shows the effectiveness of the cascaded network.}
	\setlength{\tabcolsep}{1mm}{
	\begin{tabular}{|l|c|c|c|c|c|c|c|c|}
		\hline
		\multicolumn{1}{|c|}{\multirow{2}{*}{Method}}                                    & \multirow{2}{*}{Resolution}                                            & \multicolumn{4}{c|}{Lower is better}                                                                                                                                                                                                                          & \multicolumn{3}{c|}{Higher is better}                                                                                                                                                         \\ \cline{3-9} 
		\multicolumn{1}{|c|}{}                                                           &                                                                        & Abs Rel                                                       & Sq Rel                                                        & RMSE                                                          & RMSE log                                                      & a\textless{}1.25                                              & a\textless{}1.25                                              & a\textless{}1.25                                              \\ \hline
		\begin{tabular}[c]{@{}l@{}}Baseline\\ CMDEN layer 2\\ CMDEN layer 3\end{tabular} & \begin{tabular}[c]{@{}c@{}}640x192\\ 640x192\\ 640x192\end{tabular}    & \begin{tabular}[c]{@{}c@{}}0.115\\ 0.113\\ \textbf{0.112}\end{tabular} & \begin{tabular}[c]{@{}c@{}}0.903\\ 0.823\\ \textbf{0.809}\end{tabular} & \begin{tabular}[c]{@{}c@{}}4.863\\ 4.774\\ \textbf{4.720}\end{tabular} & \begin{tabular}[c]{@{}c@{}}0.193\\ 0.192\\ \textbf{0.191}\end{tabular} & \begin{tabular}[c]{@{}c@{}}0.877\\ 0.877\\ \textbf{0.878}\end{tabular} & \begin{tabular}[c]{@{}c@{}}0.959\\ 0.959\\ \textbf{0.960}\end{tabular} & \begin{tabular}[c]{@{}c@{}}\textbf{0.981}\\ \textbf{0.981}\\ \textbf{0.981}\end{tabular} \\ \hline
		\begin{tabular}[c]{@{}l@{}}Baseline\\ CMDEN layer 2\\ CMDEN layer 3\end{tabular} & \begin{tabular}[c]{@{}c@{}}1024x320\\ 1024x320\\ 1024x320\end{tabular} & \begin{tabular}[c]{@{}c@{}}0.115\\ 0.113\\ \textbf{0.110}\end{tabular} & \begin{tabular}[c]{@{}c@{}}0.882\\ 0.808\\ \textbf{0.794}\end{tabular} & \begin{tabular}[c]{@{}c@{}}4.701\\ 4.648\\ \textbf{4.591}\end{tabular} & \begin{tabular}[c]{@{}c@{}}0.190\\ 0.189\\ \textbf{0.188}\end{tabular} & \begin{tabular}[c]{@{}c@{}}\textbf{0.879}\\ \textbf{0.879}\\ \textbf{0.879}\end{tabular} & \begin{tabular}[c]{@{}c@{}}0.961\\ 0.961\\ \textbf{0.962}\end{tabular} & \begin{tabular}[c]{@{}c@{}}\textbf{0.982}\\ \textbf{0.982}\\ \textbf{0.982}\end{tabular} \\ \hline
	\end{tabular}}
\end{table}

\section{Conclusion}
In this work, we show how to divide the scene into various parts of different sight distances, and then use the cascaded network for model training. We have developed a new masking approach to avoid the mutual influence between the network layers during training. In the context of self-supervised learning, we analyzed the relationship between monocular depth estimation and sight distance. We show superior performance on the KITTI Eigen split, exceeding all baselines. We also used an ablation study to test the cascaded network's effectiveness when the number of network layers is different. The results show that our cascading strategy is significant, and the number of layers can be increased or decreased according to actual needs.

\section{Acknowledgements}
This research was supported by School of Computer and Information Engineering, Zhejiang Gongshang University.

\section*{References}

\bibliography{mybibfile}

\end{document}